\documentclass[11pt,a4paper]{article}
\usepackage[hyperref]{emnlp-ijcnlp-2019}
\usepackage{times}
\usepackage{latexsym}

\usepackage{url}
\usepackage{graphicx}
\usepackage{caption}
\usepackage{subcaption}
\usepackage{hyperref}
\aclfinalcopy 

\title{Reasoning Over Paths via Knowledge Base Completion}

\author{Saatviga Sudhahar, Ian Roberts and Andrea Pierleoni \\
  Healx, Cambridge, UK\\
  {\tt \{saatviga.sudhahar, ian.roberts, andrea.pierleoni\}@healx.io} \\
  }

\date{}

\begin{document}
\maketitle
\begin{abstract}
Reasoning over paths in large scale knowledge  graphs  is  an  important  problem  for many applications.   In this paper we discuss  a  simple  approach  to  automatically build  and  rank  paths  between  a  source and target entity pair with learned embeddings using a knowledge base completion model (KBC). We assembled a knowledge graph by mining the available biomedical scientific literature and extracted a set of high frequency paths to use for validation. We demonstrate that our method is able to effectively rank a list of known paths between a pair of entities and also come up with plausible paths that are not present in the knowledge graph. For a given entity pair we are able to  reconstruct the highest ranking path 60\% of the time within the top 10 ranked paths and achieve 49\% mean average precision. Our approach is compositional since any KBC model that can produce vector representations of entities can be used.
\end{abstract}

\section{Introduction}
A large amount of work has been dedicated in the past on Knowledge base completion (KBC) which is an automated process that adds missing facts that can be predicted from existing ones already in the Knowledge base. This is crucial for the use of large Knowledge bases in many downstream applications. However explaining the predictions given by a KBC algorithm is quite important for several real world use cases. For example in recommender systems, a knowledge graph of users, items and their interactions are used to recommend an item to a user based on the user’s interactions on several items. The ability to explain and reason on the decision is of critical importance to add knowledge to recommender systems. Similarly in a knowledge graph consisting human biological data such as genes, drugs, symptoms and diseases, it is crucial to know which gene and symptoms were involved in predicting a drug for a disease. This requires automatic extraction and ranking of multi-hop paths between a given source and a target entity from a knowledge graph.


Previous work has focused on using path information in knowledge graphs for KBC known as path-based inference \cite{Lao:2011:RWI:2145432.2145494,gardner-etal-2014-incorporating,neelakantan-etal-2015-compositional,das-etal-2017-chains}, in which a model is trained to predict missing links between a given pair of entities taking as input several paths that existed between them. Paths are ranked according to a scoring method and used as features to train the model. Embedding-based inference models \cite{bordes2013translating,lin2015learning,nickel2011three,socher2013reasoning, trouillon2016complex} for KBC learn entity and relation embeddings by solving an optimization problem that maximises the plausibility of known facts in the knowledge graph. A third set of models bridge path-based and embedding-based inference with deep-reinforcement learning for reasoning in knowledge graphs \cite{xiong2017deeppath,das2017go,lin2018multi,song2019explainable}. All these models address various objectives. They try to predict missing links in a graph by ranking the target entities given a query entity, infer new relations between entity pairs given a set of multi-hop input paths between those pairs or infer target entities while also reasoning on paths identified. Our work is inline with the third objective in which we propose a simple approach that combines embedding-based models with a path building and ranking strategy to come up with the most probable explanations for a given prediction from a source to target. The problem can be formulated as follows: Given a source entity $e_1$ and target entity $e_2$ the goal is to first come up with a set of meaningful paths $P(e_1,e_2)=\{p_1,p_2....p_n\}$ connecting $e_1$ and $e_2$. $P(e_1,e_2)$ can contain known and predicted edges as given by the embedding-based model. Ranking of the paths is given by,

\begin{equation}
    R_{e_1e_2}=f_\Theta(e_1,e_2|P(e_1,e_2))
\end{equation}
where $f_\Theta$ denotes the underlying model with parameters $\Theta$, and $R_{e_1e_2}$ presents the ranking of the paths. 

\begin{table}[b]
\begin{center}
\begin{tabular}{lllll}
\hline  \textbf{Gene} &  \textbf{Phenotype} &  \textbf{Disease} \\ 
\hline
CA1 & Hypothermia &  Ischemia\\
CA1  &  Neuronal loss & Ischemia\\
CA1 &  Hyperglycemia &  Ischemia\\
\hline  \textbf{Gene}  &  \textbf{Drug} &  \textbf{Disease} \\ 
\hline 
BTK  & Ibrutinib & Chronic  \\
 & & lymphocytic leukemia \\
BTK  & Acalabrutinib &  Chronic  \\
& & lymphocytic leukemia \\
\hline
\end{tabular}
\end{center}
\caption{\label{font-table} Example paths for query types `\textit{Gene-Phenotype-Disease}', `\textit{Gene-Drug-Disease}' showing the ability to reason over 1-hop. }
\end{table}

Table 1 shows two examples of ranked 1-hop paths for query types,  `\textit{Gene-Phenotype-Disease}' and `\textit{Gene-Drug-Disease}'. In the first example, given a predicted fact that Gene `Carbonic Anhydrase 1 (CA1)' is associated with Disease `Ischemia', the most probable explanations can be generated by building such paths between the entities. Gene `CA1' is linked to Phenotypes `Hypothermia', `Neuronal loss' and `Hyperglycemia' and these Phenotypes are linked with Disease `Ischemia' and therefore reasoning on the fact that Gene `CA1' is associated with Disease `Ischemia'. Similarly in second example Gene `Bruton tyrosine kinase (BTK)' is associated with Disease `Chronic lymphocytic leukemia' since the Drugs `Ibrutinib' and `Acalabrutinib' are found to be linked with Gene `BTK' and the Disease. Our method is able to extract such paths and rank them providing the ability to reason on predictions.

From a knowledge graph data set, we initially train an embedding-based KBC model that can predict target entities given a source, relation pair. The KBC model is then used to build a set of 1-hop paths between a given source and target entity and paths are ranked according to a scoring function. Any embedding based KBC model capable of producing separate vector representations of entities can be used for training in our method. Our approach is quite suitable for downstream applications since it only requires training the model once with the whole data set and does not involve any additional training overhead. We discuss two main experiments in the paper, Experiment 1 showing the model's ability to reconstruct the highest ranking path in the set of retrieved ranked paths in spite of not seeing it during training and Experiment 2 that proves the ranking capability alone of the trained model with known paths. We show that our method is able to reconstruct the highest ranking path 60\% of the time in the top 10 ranked paths for a given set of entity pairs also achieving 49\% mean average precision. There is almost a 3-fold increase in the ranking correlation between predicted ranks and ground truth ranks of a longer list of known paths when compared to random. To our knowledge this is the first paper that is focused on trying to use path ranking to identify relevant entities bridging a pair of known entities and therefore not directly comparable with other approaches.

The paper is organised as follows: Section 2 discusses related work in this area; Section 3 shows details of how the data sets of triples and paths between entity pairs were generated and elaborates on model path construction and path scoring for ranking the paths; Section 4 discusses the statistics of the data sets; Section 5 shows and discusses experimental results on the model's ability to recover paths and ranking in experiment 1 and ranking alone of known paths in experiment 2. Section 6 concludes the work and discusses possible future directions.

\section{Related Work}
One of the first path building approach in literature is the Path-Ranking Algorithm (PRA) \cite{lao2010fast,gardner-etal-2013-improving,gardner-etal-2014-incorporating} which uses random walk with restarts to rank entities in a graph relative to a source entity. Random walks constrained to follow a set of edge types are performed in the graph to produce relational feature weights. The weights are combined to predict the probability that a relation between a source, target entity pair using a per-target relation binary classifier. The bottleneck of this method is the explosion of the feature space when there are millions of distinct paths obtained from random exploration and the fact that it relies on a binary classifier per target relation which is not scalable. 

\newcite{guu-etal-2015-traversing} proposed a method to build relation paths between entity pairs using compositional techniques to address knowledge base completion and path query answering problems. They learn representations of nodes and relations by adapting the scoring functions from pre-existing KBC models such as TransE \cite{bordes2013translating} and RESCAL \cite{nickel2011three}. They cannot use models in which the scoring function cannot be decomposed in a way to produce separate vectors for a source/relation without the target. In contrast our method can use any embedding based KBC model for training. Another limitation in their work is that they model only a single path between an entity pair but we model multiple paths. \newcite{toutanova2015representing} propose a similar framework and also model intermediate entities in the path, training a convolutional neural network to rank a set of target entities given a source and relation pair. Path-RNN \cite{neelakantan-etal-2015-compositional} is also a compositional model that takes paths between entity pairs as input and infers new relations between them. However it uses only the relation information in the path and ignores modelling intermediate entities, and it requires training a model for each relation type making it not usable in downstream applications. \newcite{das-etal-2017-chains} train a high capcity RNN model to infer relations between entity pairs by taking several multi-hop paths between them as input and score pool them using techniques such as averaging across all paths or the top-$k$ paths.

Recent work in explainable reasoning is used in recommender systems. \newcite{wang2019explainable} propose a knowledge-aware path recurrent network to generate representation for each path by composing semantics of entities and relations and reason on paths to infer user preferences for items. Pre-defined meta-paths and LSTMs have been used to model the paths between users and items via breadth-first-search (BFS). Practically this is inefficient and can miss many meaningful paths. To address these challenges, \newcite{song2019explainable} proposed a deep reinforcement learning based method that can automatically generate meaningful paths between a user and item via policy gradient methods. What we present in this paper is a very simple approach to come up with a set of most relevant paths consisting of known and predicted links between a given source and target entity in a knowledge graph using embedding based models. Its also possible to extend this approach to use contextualised knowledge graph embedding models \cite{dolores} which learn representations based on an entire path in the graph rather than a triple. We believe using this in conjunction with our method in turn could produce more meaningful paths in the path building process. This is left for future work.

\section{Methodology}

\subsection{Generating the Data}
In this section we discuss how the data set on which we perform the experiments was generated. This includes extraction of triples from the Pubmed abstracts data set using a NLP pipeline and extraction of 1-hop paths from the triple data set. The 1-hop path data set is used to construct a ranked list of paths for 100 entity pairs to generate ground truth data for the path ranking experiments. 

\subsubsection{Extracting triples}
We used the output of the Open Targets Library NLP pipeline\footnote{ https://github.com/opentargets/library-beam} to access all the subject-verb-object (SVO) triples from the  biomedical abstracts released by PubMed and updated to March 2019, for a total of more than 29M documents and 540M SVO triples. From this set we selected only the SVO triples between entities linked to a specific biomedical ID, discarding the ones with `concept' entity type and stop words. From the filtered list we built an undirected weighted knowledge graph linking each pair of entities identified by a triple with a weight representing the number of documents in which the triple was found. 

\subsubsection{Extracting 1-hop paths}
While it is possible to access many knowledge bases like the one described above in graph format, to our knowledge there is currently none focused on building relations spanning more than two entities. We leveraged the ability to track SVO triples occurring in the same abstract and built a data set of paths spanning three entities by making the assumption that if $A$ is related to $B$, and $B$ is related to $C$ in the same abstract, then its safe to assume that $A$ is related to $C$ via $B$. By counting the number of occurrences of such connected triples across different documents we are also able to weight the paths,  and thus can generate a ranked lists of paths starting from source entity $A$ to target entity $C$ by scoring higher the most frequent paths. 

By combining several 1-hop paths, it is possible to also build multi-hop paths described within the same document however, in this data set, even the 2-hop paths are too rare to be used for a test of statistical significance.

\subsection{Model Path Construction}
The primary objective of our method is to automatically build a set of paths between a given source and target entity pair in a knowledge graph and rank them using a scoring method to come up with the most relevant paths connecting the pair. We first train a 200 dimensional node, relation embedding model with the data set using ProjE. ProjE is a neural network based KBC model that fills the missing information in a knowledge graph by learning joint embeddings of the graph's entities and links. Given an input triple $\langle h,r,? \rangle$ the model can find the optimal ordered list of tail entities and vice versa. 

Let $E$ be the set of all entities and $R$ the set of all relation types in the data set. For a given entity pair ($e_1$, $e_2$), we choose the set of suitable relation types $r_t$ $\subset$ $R$ to be used and query the trained model with ($e_1$, $r_t$) and ($e_2$, $r_t$) pairs. The set of suitable relations types are the ones that agree with the query type. For entity pair with types, \textit{`Drug:Disease'}, the corresponding query types are \textit{`Drug-Gene-Disease }' or \textit{`Drug-Phenotype-Disease'} in our data set. Therefore the method will choose \textit{`Drug\_Gene'}, \textit{`Drug\_Phenotype'}, \textit{`Disease\_Gene'}, \textit{`Disease\_Phenotype'} relations to perform the predictions. This results in a set of top $n$ predicted tail and head entities $e_t$, $e_h$ $\subset$ $E$ for each relation type. If the same entity was predicted again for $e_1$ or $e_2$ but with another relation type from previous, we retain the relation type producing the higher prediction rank in the graph. We create an undirected graph $G$ out of $e_1$, $e_2$, and the predicted sets of entities $e_t$, $e_h$ with their relations. From this graph we extract all 1-hop paths connecting $e_1$ and $e_2$. 

The same approach can be used iteratively to build multi-hop paths between a pair of entities. Given the limits of the available dataset, in this paper we limit the discussion to the 1-hop case.

\subsection{Path Scoring}
Once the paths are extracted a scoring function is used to score them for ranking. A path $p$ consists of a source, target and an intermediate predicted entity connected by relevant relations. For 1-hop paths there are two $\langle h,t,r \rangle$ pairs along the path. The score $S_p$ for each path $p$ in the extracted set of paths $P=\{p_1, p_2...p_n\}$ is given by,

1. Embed+predrank: sum of the prediction ranks ($R$) of the predicted entity in $p$. Paths with lowest score are the high ranking ones.
\begin{equation}
    S_p = \sum_{h,t,r \in p}{R}
\end{equation}

2. Embed+cosine: sum of the cosine similarities between entity embeddings in $p$. The most relevant paths receive higher scores. 
\begin{equation}
    S_p = \sum_{h,t,r \in p}{cos(e_h,e_t)}
\end{equation}

Embed+predrank score only considers the rank of the predicted entities as given by the KBC model while Embed+cosine score uses the actual learned embedding to score the paths which is more informative than the previous. In the next section we show in detail the data set used and the experiments performed proving the significance of the path ranking method in two different settings. 

\section{Data}
The data set used for the experiments are constructed by extracting relations using a NLP pipeline from the Pubmed repository as shown in Section 3.1.1. It consists of 3,025,541 head, rel, tail triples. Details of this base data set is shown in Table 2. From this data set we created a set of 100 entity pairs with a ranked list of several one-hop paths $P$ connecting entity pairs ($e_1,e_2$) in the knowledge graph. 

We use two different path data sets for experiments 1 and 2. The details of this data set is shown in Table 3. Paths extracted are from query types: \textit{`Gene-Drug-Disease'}, \textit{`Gene-Phenotype-Disease'}, \textit{`Drug-Phenotype-Disease'}, \textit{`Anatomy-Phenotype-Disease'}, \textit{`Disease-Drug-Anatomy'} and \textit{`Anatomy-Gene-Organism'}. Selection of query types are purely based on simple and meaningful queries. There are at least 2 paths for each entity pair for experiment 1 and 10 for experiment 2 and they were observed twice or more in the data. Ranking of paths for each entity pair are based on how frequent they were observed in the underlying data set. We assume most frequently seen paths are reliable and should rank higher than others. We use this ranking as ground truth. An archive containing the base data sets with train/test split and ground truth path ranking data sets for experiment 1 and 2 presented are available here.\footnote{https://storage.googleapis.com/pubmed-path-kg/pubmed\_exp\_data.zip}
\begin{table}[t]
\begin{center}
\begin{tabular}{lrl}
\hline  & \# &\\ \hline
triples & 3,025,541 &\\
entities & 38,772 &\\
entity types &  10 & \\
relation types & 54 &\\
1-hop paths & 5.5M &\\
\hline
\end{tabular}
\end{center}
\caption{\label{font-table} Triple Data set statistics. }
\end{table}
\begin{table}[t]
\begin{center}
\begin{tabular}{lll}
\hline &  Exp 1 &  Exp 2 \\ 
\hline
query types & 4 & 6 \\
avg no of paths & 6 & 33 \\
min no of paths & 2  & 10\\
total paths & 573 & 3356\\
\hline
\end{tabular}
\end{center}
\caption{\label{font-table} Path data set statistics used in experiment 1 and 2. }
\end{table}
\begin{table*}[t]
\centering
\begin{tabular}{llllllllll}
  \# Pred & Scoring & Hits &MAP & Hits &MAP & Hits &MAP & Hits &MAP\\
  & &@100 &@100 &@25 &@25 &@10 &@10 &@1 &@1  \\
  \hline
  100 & Embed+predrank & 0.36 &0.13 & 0.22 &0.14 &0.1 &0.15 & 0 &0.07\\
      & Embed+cosine & 0.36 &0.27 & 0.27 &0.31 & 0.23 &0.34 & 0.07 &0.26\\
      & Baseline & 0.01 &0 & 0 &0 &0 &0 &0 &0\\
  \hline
    300 & Embed+predrank & 0.59 &0.11 & 0.22 &0.14 &0.1 &0.15 & 0 &0.06 \\
        & Embed+cosine & 0.86 &0.38 & 0.64 &0.45 & 0.57 &0.49 & 0.26 &0.47\\
      & Baseline & 0.01 &0 & 0 &0 &0 &0 &0 &0\\
    \hline
    500 & Embed+predrank & 0.59 &0.11 & 0.22 &0.14 & 0.1 &0.15 & 0 &0.06 \\
        & Embed+cosine & 0.89 &0.37 & 0.67 &0.43 & 0.6 &0.47 & 0.26 &0.43\\
      & Baseline & 0.01 &0 & 0 &0 &0 &0 &0 &0\\
    \hline
\end{tabular}
\caption{Hits@n and MAP@n scores for Embed+predrank, Embed+cosine scoring functions comparing to baseline for top 100, 300, 500 predictions given by KBC model.
  }
\end{table*}
\section{Experimental Results}
\subsection{Path recovery and ranking}
In this experiment we show that the trained KBC model is able to recover the highest ranking path in the ground truth among the ranked set of paths retrieved by our ranking method for entity pairs in spite of not seeing  the links belonging to the path in the training set. We also report on the mean average precision (MAP) obtained for top $n$ number of paths retrieved by the model. From the ground truth data set we choose one triple in the top ranking path $p_1$ for each entity pair and remove that triple from the training set. For example, if the highest ranking path is,
\textit{`Drug1-Gene1-Disease1'} for entity pair \textit{`Drug1:Disease1'}, we remove one triple from the path which could be \textit{`Gene1,Disease1,Gene1\_Disease1'} from the training set. We train an embedding model with ProjE in the resulting data set using the following parameters: \textit{d} 200, \textit{epochs} 200, \textit{lr} 0.0001, \textit{batchsize} 128, \textit{dropout} 0.5, \textit{lossweight} 0.0001. We use a negative sampling rate of 0.25 and minimise the loss using \textit{adam} optimiser. 

The trained model is then used to build a set of ranked paths between the source and target entity pairs as described in section 3.2. What we test here is whether the model is able to recover the highest ranking path $p_1$ from the ground truth for each entity pair in the ranked list of paths. Table 4 shows Hits@n (1,10,25,100) which is the number of times $p_1$ was ranked in the top $n$ ranked paths out of all entity pairs for scoring functions Embed+predrank and Embed+cosine and reports MAP for the $n$ ranked paths retrieved. We compare the result with a baseline in which we build paths by choosing the most similar entity to the source and target entities by cosine similarity and rank them based on the total score obtained. Though it still uses the learned embedding to compute similarities it does not consider any predictions made by the model. Column 1 refers to the number of top $n$ predictions (100,300 and 500) that we request from the KBC model. 

According to results in Table 4, Embed+cosine scoring function yields better Hits@n and MAP@n scores than Embed+pred rank in general. When more predictions are requested from the KBC model, we observe an increase in the Embed+cosine scores for both Hits@n and MAP@n. Increasing the number of predictions in the KBC model, allows for more entities to be predicted which in turn increases the depth of path search, therefore leading to a better chance that $p_1$ could be retrieved for Hits@n and more correct paths for MAP@n. When using the top 500 predictions, the model is able to recover $p_1$ on top 26\% of the time, within the top 10, 60\% of the time and within the top 25, 67\% of the time. The scores are even higher when more ranked paths are extracted according to Hits@100. Within 100 and 300 predictions the model is able to achieve 34\% and 49\% MAP@10 and 31\% and 45\% MAP@25 using Embed+cosine scoring. When compared to the baseline we see the scores for both Hits@n and MAP@n obtained by our path building and ranking method is quite significant. Note that the baseline does not depend on the number of predictions requested by the model and hence it remains same for all three settings.

\subsection{Path ranking}
In this experiment we want to test how good the model can rank known paths, that consists of known links between entities in a graph. In order to do this we retrieve the rank for known paths in the ground truth as given by the model. We train a KBC model with ProjE again, but this time the model is trained on the whole data set. For each entity pair in the same test set, the model produces a set of ranked paths from which we compute the relative ranking of the known paths in the ground truth. For example if entity pair ($e_1$, $e_2$) has 15 ranked paths, and our model produced 25 ranked paths in total, we compute the relative ranking of only the 15 paths for this test. Our test statistic is spearman ranking correlation between the ranks retrieved by the model and the ground truth. The same test is repeated after randomly permuting the paths in the ground truth. We also compute the ranking correlation between the randomly permuted set and the actual ground truth. The ground truth data set used for this experiment has at least 10 ranked paths for each entity pair and on average has 33 paths per pair as shown in Table 3. We increased the number of predictions given by the KBC model to 1500 in this experiment because it increases the depth of path search and retrieves all known paths in the ground truth. This was purely done for evaluation purposes since its important to compare the rankings of all paths in the ground truth. 

Table 5 shows the Spearman ranking correlation ($r_s$) for $R_p$-$R_t$ ranks of paths in ground truth ($R_t$) and their relative ranking in predicted paths ($R_p$), $R_p$-$R_r$ ranks of randomly permuted paths ($R_r$) and their relative ranking in predicted paths ($R_p$), $R_t$-$R_r$ ranks of paths in ground truth ($R_t$) and their relative ranking in randomly permuted paths ($R_r$). We show $r_s$ observed for all entity pairs (100) in row 1 and for the ones with at least 20 or more paths (67) in row 2. $r_s$ decreases with ranking complexity as we can see but in both cases $r_s$ with the ground truth is better compared to $r_s$ with the permuted set showing almost 3-fold (2.8) increase in the most difficult case. This proves the significance of the model's ranking capability.

\begin{table}[t]
\centering
\begin{tabular}{llll}
  \# Paths & Coverage & Rank pair & $r_s$ \\
  \hline
  $\geq$10 & 100 & $R_p$-$R_t$ & 0.45 \\
         & & $R_p$-$R_r$ & 0.24\\    
         & & $R_t$-$R_r$ & 0.26 \\
    \hline
      $\geq$ 20 & 67  & $R_p$-$R_t$ & 0.37 \\
         & & $R_p$-$R_r$ & 0.13\\    
         & & $R_t$-$R_r$ &  0.12\\
    \hline
\end{tabular}
\caption{Spearman ranking correlation ($r_s$) for $R_p$-$R_t$ ranks of paths in ground truth and their relative ranking in predicted paths, $R_p$-$R_r$ ranks of randomly permuted paths and their relative ranking in predicted paths, $R_t$-$R_r$ ranks of paths in ground truth and their relative ranking in randomly permuted paths.
  }
\end{table}

\begin{figure}[h]
     \centering
     \begin{subfigure}[b]{0.5\textwidth}
         \centering
         \includegraphics[width=5cm]{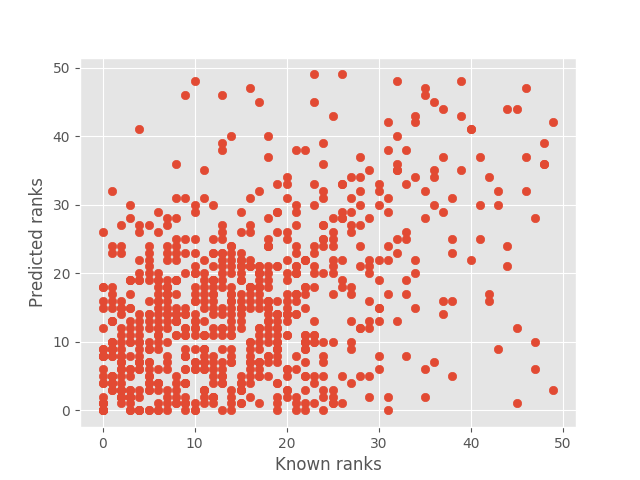}
         \caption{}
     \end{subfigure}
     \begin{subfigure}[b]{0.5\textwidth}
         \centering
         \includegraphics[width=5cm]{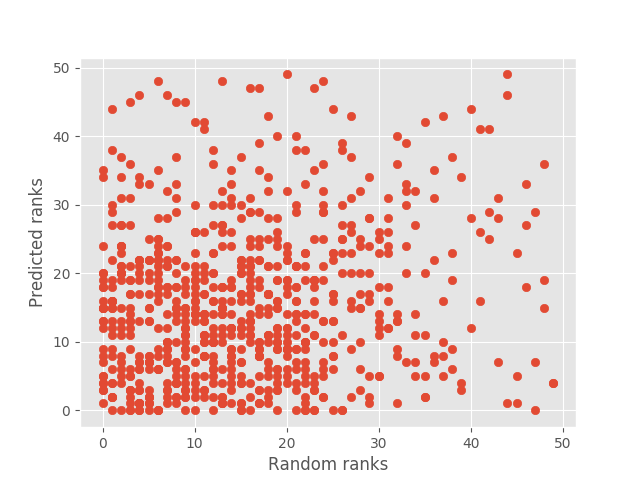}
         \caption{}
     \end{subfigure}
\caption{(a) Ranks of known paths in ground truth vs their relative rank in predicted paths, (b) Ranks of randomly permuted paths in the ground truth vs their relative rank in predicted paths for all entity pairs with at least 20 or more paths.}
\end{figure} 

The ground truth data set used for this experiment is quite scarce and not very well distributed in terms of ranking. Since ranks are generated based on frequency of the paths observed, paths with same frequency get similar rankings. This is a limitation in the data set. We expect the method to perform better in cases where there exists larger number of paths with an evenly distributed ranking. This is left for future work.  

Figure 1a plots the ranks of known paths in ground truth vs their relative rank in predicted paths and Figure 1b plots the ranks of randomly permuted paths in the ground truth vs their relative rank in predicted paths for all entity pairs with at least 20 or more paths reported in row2 of Table 5. Points in Figure 1a are mostly correlated and shows a huge tail towards the bottom left indicating the highest ranking paths are well correlated between known and predicted ranks. Figure 1b is quite scattered compared to 1a as indicated by results in Table 5. 


\section{Conclusion \& Future Work}
In this paper we proposed a simple method to automatically build and rank paths between a source and target entity pair using embedding based knowledge base completion (KBC) models. To our knowledge this is the first paper that is focused on trying to use path ranking to identify relevant entities bridging a pair of known entities and hence not directly comparable with other approaches. To this purpose we built a data set that allow us to test our hypothesis. We demonstrated that our method is able to effectively rank known paths if available and also infer important missing links between entities during the path building process, ranking them high when they are significant. Any embedding based KBC model can be used after initial training allowing for more flexibility and also less overhead. The number of paths built between a source and target is dependent on the number of predictions requested from the KBC model as the search space for paths increase with this parameter. 

As future work we plan to build a benchmark data set for multi-hop paths with a smooth distribution of ranked paths, evaluate the same approach and also perform an extensive evaluation with other state-of-the-art KBC models based on convolutional networks and matrix factorisation for path construction and ranking. We intend to also work on extending our method to use contextualised embedding representations to make better use of path information in the graph which we believe will have a positive impact on the current path building and ranking methodology.

\bibliography{emnlp-ijcnlp-2019}
\bibliographystyle{acl_natbib}

\appendix

\end{document}